\documentclass[journal]{IEEEtran}

\usepackage{graphicx}
\usepackage{epstopdf}
\usepackage{subcaption}
\usepackage[table,xcdraw]{xcolor}
\usepackage{hyperref}
\usepackage{times}
\usepackage{amsmath}
\usepackage{amssymb}
\usepackage{array}
\usepackage{overpic}
\usepackage{float}
\usepackage[ruled,linesnumbered]{algorithm2e}
\usepackage{enumitem}
\usepackage{booktabs}
\usepackage[english]{babel}
\usepackage{amsthm}
\usepackage{svg}
\usepackage{adjustbox}

\graphicspath{ {picture/} }

\begin{document}

\title{Adaptive 3D Gaussian Splatting Video Streaming}

\author{Han Gong,
        Qiyue Li,~\IEEEmembership{Senior Member,~IEEE},
        Zhi Liu,~\IEEEmembership{Senior Member,~IEEE},
        Hao Zhou,~\IEEEmembership{Member,~IEEE},
        Peng Yuan Zhou,~\IEEEmembership{Member,~IEEE},
        Zhu Li,~\IEEEmembership{Senior Member,~IEEE},
        Jie Li,~\IEEEmembership{Member,~IEEE}
        	
\thanks{Han Gong and Qiyue Li are with the School of Electrical Engineering and Automation, Hefei University of Technology, and Engineering Technology Research Center of Industrial Automation Anhui Province, Hefei, Anhui 230009, China. (e-mail: han\_gong@mail.hfut.edu.cn; liqiyue@mail.ustc.edu.cn).}

\thanks{Zhi Liu is with The University of Electro-Communications, Tokyo, Japan (e-mail: liuzhi@uec.ac.jp).}

\thanks{Hao Zhou is with the School of Computer Science and Technology, University of Science and Technology of China, Hefei, China (e-mail: kitewind@ustc.edu.cn).}

\thanks{Peng Yuan Zhou is with the Department of Electrical and Computer Engineering, Aarhus University, Aarhus, Denmark. (e-mail: pengyuan.zhou@ece.au.dk).}

\thanks{Zhu Li is with the Department of Computer Science and Electrical Engineering, University of Missouri–Kansas City, Kansas, USA (e-mail: lizhu@umkc.edu).}
\thanks{Jie Li is with the School of Computer Science and Information Engineering, Hefei University of Technology, Hefei, China (e-mail: lijie@hfut.edu.cn).}

\thanks{Corresponding author: Jie Li.}
}

\maketitle

\begin{abstract}

The advent of 3D Gaussian splatting (3DGS) has significantly enhanced the quality of volumetric video representation. Meanwhile, in contrast to conventional volumetric video, 3DGS video poses significant challenges for streaming due to its substantially larger data volume and the heightened complexity involved in compression and transmission. To address these issues, we introduce an innovative framework for 3DGS volumetric video streaming. Specifically, we design a 3DGS video construction method based on the Gaussian deformation field. By employing hybrid saliency tiling and differentiated quality modeling of 3DGS video, we achieve efficient data compression and adaptation to bandwidth fluctuations while ensuring high transmission quality. Then we build a complete 3DGS video streaming system and validate the transmission performance. Through experimental evaluation, our method demonstrated superiority over existing approaches in various aspects, including video quality, compression effectiveness, and transmission rate.

\end{abstract}


\IEEEpeerreviewmaketitle

\section{Introduction}

Volumetric video, offering users an immersive 6-DoF (six degrees of freedom) viewing experience, is a key application in the 6G era and holds significant development potential.
The evolution of volumetric video can be broadly categorized into two paradigms: explicit representation schemes and implicit representation schemes.
Traditional explicit representation schemes such as point cloud, voxel, and mesh, often struggle with accurately depicting textures, and the fixed RGB values constrain their capacity to render light and shadow effectively. In contrast, implicit representation schemes, exhibit highly realistic scene rendering capabilities. For example, NeRF’s volume rendering approach excels in capturing intricate scene details and light propagation with remarkable precision. However, NeRF relies on neural networks to query 3D scene features, lacking an explicit model for interactive editing. This absence of a clear correspondence between the visual output and its underlying representation poses significant challenges for downstream tasks such as gaming or remote video communication that require model editing \cite{zhou2024feature}.

As an innovative explicit scene representation paradigm, 3D Gaussian Splatting (3DGS) constructs 3D environments using hundreds of thousands of anisotropic Gaussian primitives—each functioning as a dynamically adjustable ellipsoid that encodes spatial coordinates, orientation and scale, opacity, and directional color properties via spherical harmonics. During real-time rendering, these Gaussian primitives are projected into 2D screen-space footprints through perspective transformation, followed by depth-sorted optical blending, where nearer primitives naturally occlude farther ones. Final images are synthesized via per-pixel radiance integration that fuses color and opacity in a differentiable manner. This pipeline enables photorealistic view synthesis while preserving explicit geometric control, marking a significant advancement over implicit neural radiance fields. Building on this explicit architecture, 3DGS models feature structured collections of Gaussian primitives in space, each encapsulating shape, color, and opacity attributes centered at its position.

3DGS videos offer significantly better visual quality than traditional explicit representations and enable easier client-side editing and interaction, unlike NeRF. This explicitness benefits applications like gaming and remote communication by allowing real-time modifications and transmission of editable 3D models, enhancing immersion and interactivity. Table \ref{tab:3d-representations} compares 3DGS with other volumetric video formats. Thus, streaming 3DGS delivers a novel, spatialized immersive experience.


\begin{table*}[ht]
\centering
\renewcommand{\arraystretch}{1.5}
\caption{Comparison of different volumetric video representations.} 
\label{tab:3d-representations} 
\begin{tabular}{|>{\centering\arraybackslash}p{2cm}|>{\centering\arraybackslash}p{1.5cm}|>{\centering\arraybackslash}p{1.5cm}|>{\centering\arraybackslash}p{1.5cm}|>{\centering\arraybackslash}p{1.5cm}|>{\centering\arraybackslash}p{1.5cm}|>{\centering\arraybackslash}p{1.5cm}|>{\centering\arraybackslash}m{2cm}|}
\hline
\rowcolor{lightgray!25} \begin{tabular}{@{}c@{}}Representation\\ Name\end{tabular} & Size   & \begin{tabular}{@{}c@{}}Visual\\ Quality\end{tabular} & \begin{tabular}{@{}c@{}}Rendering\\ Cost\end{tabular} & Editability & \begin{tabular}{@{}c@{}}Salience\\ Complexity\end{tabular} & \begin{tabular}{@{}c@{}}Representation\\ Format\end{tabular} & Example \\ \hline
\rowcolor{blue!10} Point Cloud    & Large  & Low  & Low  & Easy  & Medium  & Explicit       & \includegraphics[width=0.065\textwidth]{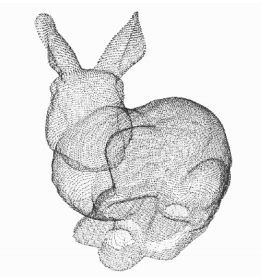} \\ \hline
\rowcolor{blue!5} Voxel          & Medium & Low  & Low  & Easy  & Easy  & Explicit           & \includegraphics[width=0.065\textwidth]{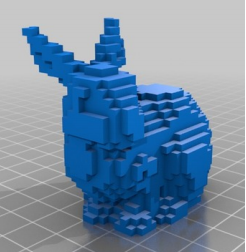} \\ \hline
\rowcolor{blue!10} Mesh           & Medium & Medium  & Medium  & Medium  & Easy  & Explicit           & \includegraphics[width=0.065\textwidth]{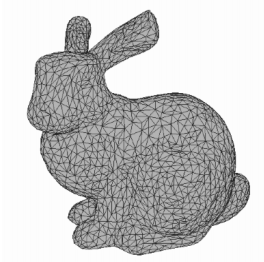} \\ \hline
\rowcolor{blue!5} NeRF           & Medium & High & High & Hard & Hard  & Implicit          & \includegraphics[width=0.065\textwidth]{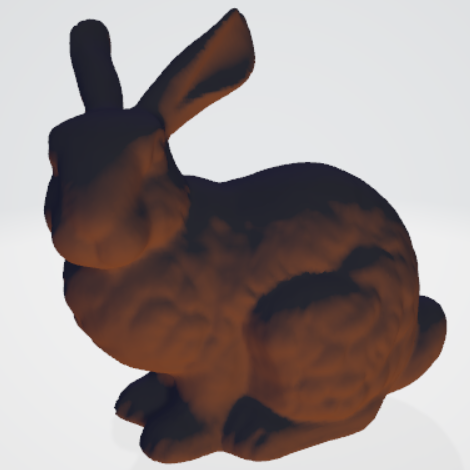} \\ \hline
\rowcolor{blue!10} 3DGS           & Huge   & High  & Medium & Easy & Hard  & Explicit          & \includegraphics[width=0.065\textwidth]{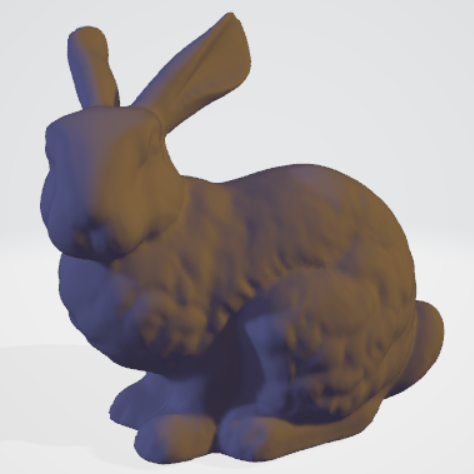} \\ \hline
\end{tabular}
\end{table*}

However, the unique characteristics of 3DGS preclude the direct application of streaming methods designed for other types of volumetric videos to 3DGS volumetric videos.
The core challenges of 3DGS video streaming can be summarized into the following three aspects:

\begin{itemize}

 \item \textbf{Huge Data Volume of 3DGS Video:} Modeling each frame of a 3DGS video independently leads to excessive data. For example, a 30 FPS 3DGS video without compression has about 49 MB per frame, requiring up to 11.76 Gbps bandwidth for streaming \cite{kerbl20233d}. Such a large volume is impractical for direct transmission, necessitating efficient compression methods.

 \item \textbf{Saliency Extraction for Disordered Gaussian Points:} Visual saliency extraction is essential for volumetric video transmission, but Gaussian primitives in 3DGS pose challenges due to their irregular spatial distribution. Each Gaussian stores 64 attributes, significantly increasing data complexity and complicating the extraction of effective features for tasks like Field of View (FoV) prediction.

 \item \textbf{Compression of Inhomogeneous Gaussian Points:} Unlike uniform points in point clouds, each Gaussian primitive represents an ellipsoid with distinct scale, opacity, and color, which differently affect the final rendering. Therefore, compression must carefully consider the significance of each Gaussian primitive, avoiding arbitrary discarding to preserve imaging quality.

\end{itemize}

To address the aforementioned challenges and fill the existing gap in the integration of 3DGS video with streaming, we propose a novel framework that leverages adaptive tiling and differentiated quality modeling. By spatially segmenting the 3DGS video into multiple adaptive tiles based on hybrid saliency, and categorizing the content within each tile into varying quality levels, we can then select the optimal tile and its corresponding quality level for transmission, tailored to the real-time bandwidth and the user's FoV. Our contributions can be summarized as follows:
\begin{itemize}

 \item We build a comprehensive 3DGS video streaming framework based on the proposed idea, and validate our approach's feasibility.
 To the best of our knowledge, this is the first complete framework for 3DGS volumetric video streaming, covering the entire process from video construction to adaptive transmission, enabling real-time viewing on VR devices.
   
 \item We propose a 3DGS video construction method using Gaussian deformation fields to represent videos within each Group of Frames (GoF). This approach preserves video quality while compressing data by segmenting GoFs and modeling their frames with separate deformation fields, reducing the complexity of long-sequence modeling.

 \item We propose an adaptive tiling method for 3DGS videos by extracting static and dynamic saliency within each GoF to create a hybrid saliency map. This guides segmentation into adaptive tiles and enables assigning quality levels based on visual importance, preserving high-saliency areas while compressing low-saliency ones to balance quality and transmission efficiency.

 \item We propose a differential quality modeling method using saliency weights and a Gaussian mask that considers both primitive properties and tile saliency. By removing small or transparent ellipsoids, we minimize quality loss while retaining more details in high-saliency regions and aggressively compressing low-saliency areas to enhance efficiency.
 
\end{itemize}


\section{The Development of Volumetric Video Streaming}

Volumetric video has rapidly advanced alongside developments in computer graphics and communication technologies, resulting in diverse representation methods. However, its large data volumes hinder seamless deployment over typical networks. Efficient streaming requires multi-quality layering, optimized codecs balancing decoding time and data size, adaptive bitrate algorithms based on user experience, and viewpoint prediction to prioritize content within the user’s FoV.

In recent years, numerous approaches have emerged to optimize conventional volumetric video streaming along these lines. However, with the advent of Neural Volumetric Videos (NVV), represented by methods like NeRF and 3DGS, traditional streaming techniques have become increasingly inadequate. NeRF’s implicit representation, lacking a concrete spatial model, precludes the use of conventional strategies such as tiling or octree partitioning. Moreover, its requirement for a fixed viewpoint to render content and its inherently slow volumetric rendering severely limit the feasibility of real-time streaming and adaptive quality stratification based on visual fidelity.

In contrast, 3DGS adopts an explicit representation closely aligned with traditional volumetric formats, enabling the reuse of methods developed for point cloud or mesh-based video—such as spatial partitioning via tiles or octrees. Additionally, its use of rasterized rendering by projecting 3D ellipsoids onto a 2D plane is significantly faster than NeRF’s volumetric rendering, eliminating rendering speed as a bottleneck for real-time NVV streaming. Thanks to its remarkable rendering efficiency, 3DGS naturally supports smooth, real-time viewpoint transitions. It is therefore poised to supersede traditional volumetric formats and emerge as the dominant paradigm for future volumetric video, with vast potential for widespread adoption.

Numerous studies have already tackled the challenges of 3DGS video streaming. Efforts such as 3DGStream \cite{sun20243dgstream}, HiCoM \cite{gao2024hicom}, DASS \cite{liu2024dynamics}, and V3 \cite{wang2024v} focus on generating efficient dynamic 3DGS sources. Compared to conservative frame-by-frame reconstructions, dynamic 3DGS significantly reduces per-frame data while shortening training durations. Although real-time reconstruction remains out of reach, streaming-style processing—training concurrently with data acquisition—offers a promising blueprint for future real-time capabilities. Moreover, frameworks like L3GS \cite{tsai2025l3gs}, PCGS \cite{chen2025pcgs}, and TGH \cite{xu2024representing} introduce schemes for fine-grained quality stratification, paving the way for adaptive transmission under fluctuating network conditions. The LTS \cite{sun2025lts} approach further explores integrating GoF- and tile-based methods from traditional volumetric streaming into the 3DGS pipeline.

As a novel approach to volumetric video representation, research on 3DGS volumetric video streaming is still in its early stages. Effectively leveraging 3DGS for the creation of high-quality volumetric video, alongside the development of methods for compressing and transmitting 3DGS content, presents significant challenges that must be addressed. The following section outlines our proposed 3DGS video streaming scheme aimed at tackling these issues.

\begin{figure*}[htb]
    \centering
    \includegraphics[width=0.9\textwidth]{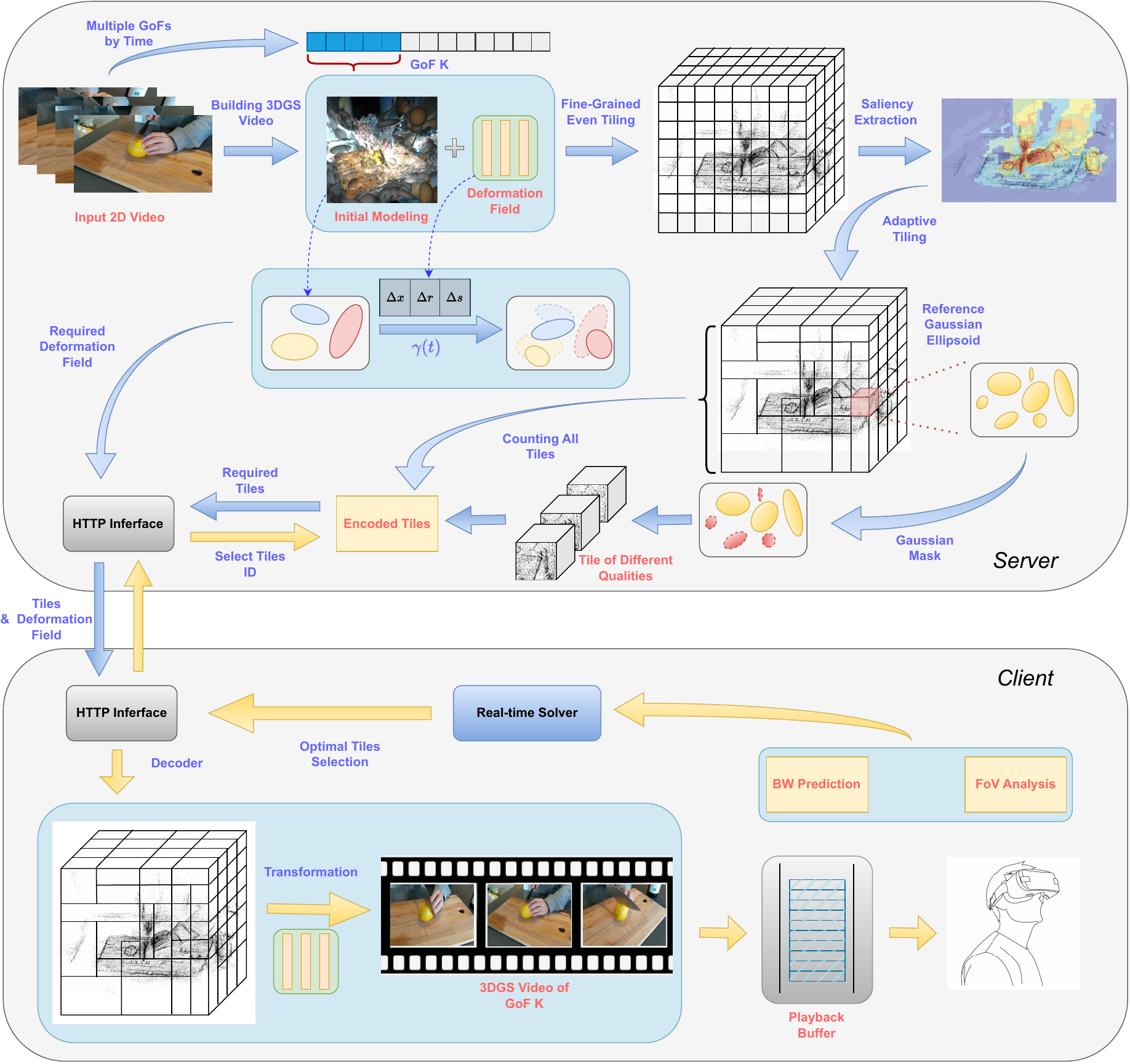}
    \caption{The overall framework of optimal 3DGS video streaming is divided into two parts: the server side and the client side.}
    \label{fig_System}
\end{figure*}

\section{3DGS Volumetric Video Streaming Framework}

This study proposes a novel streaming framework specifically designed for 3DGS video. Unlike prior 3DGS streaming research primarily focused on reconstructing streamable 3DGS video sources, our work centers on optimizing the transmission efficiency of existing 3DGS video content. Figure \ref{fig_System} illustrates the server-side architecture of our 3DGS streaming system, encompassing all processing stages from source preparation to optimized delivery, thereby providing a practical foundation for real-world 3DGS video applications. The system architecture comprises four core functional stages: (1) Gaussian Deformation Field-based Dynamic Video Construction, (2) Hybrid Saliency-driven Adaptive Tiling, (3) Differentiated Quality Modeling, and (4) QoE-driven Adaptive Transmission Optimization.

The framework first segments videos into GoFs and constructs intra-GoF deformation fields for compression; then performs spatiotemporal saliency fusion to generate adaptive tiles; subsequently establishes multi-tier quality levels per tile via Gaussian masking; and finally dynamically selects optimal tile-quality pairs under bandwidth constraints to maximize QoE. The details of each step are explained in the following section.

\begin{figure*}[htb]
    \centering
    \includegraphics[width=0.9\textwidth]{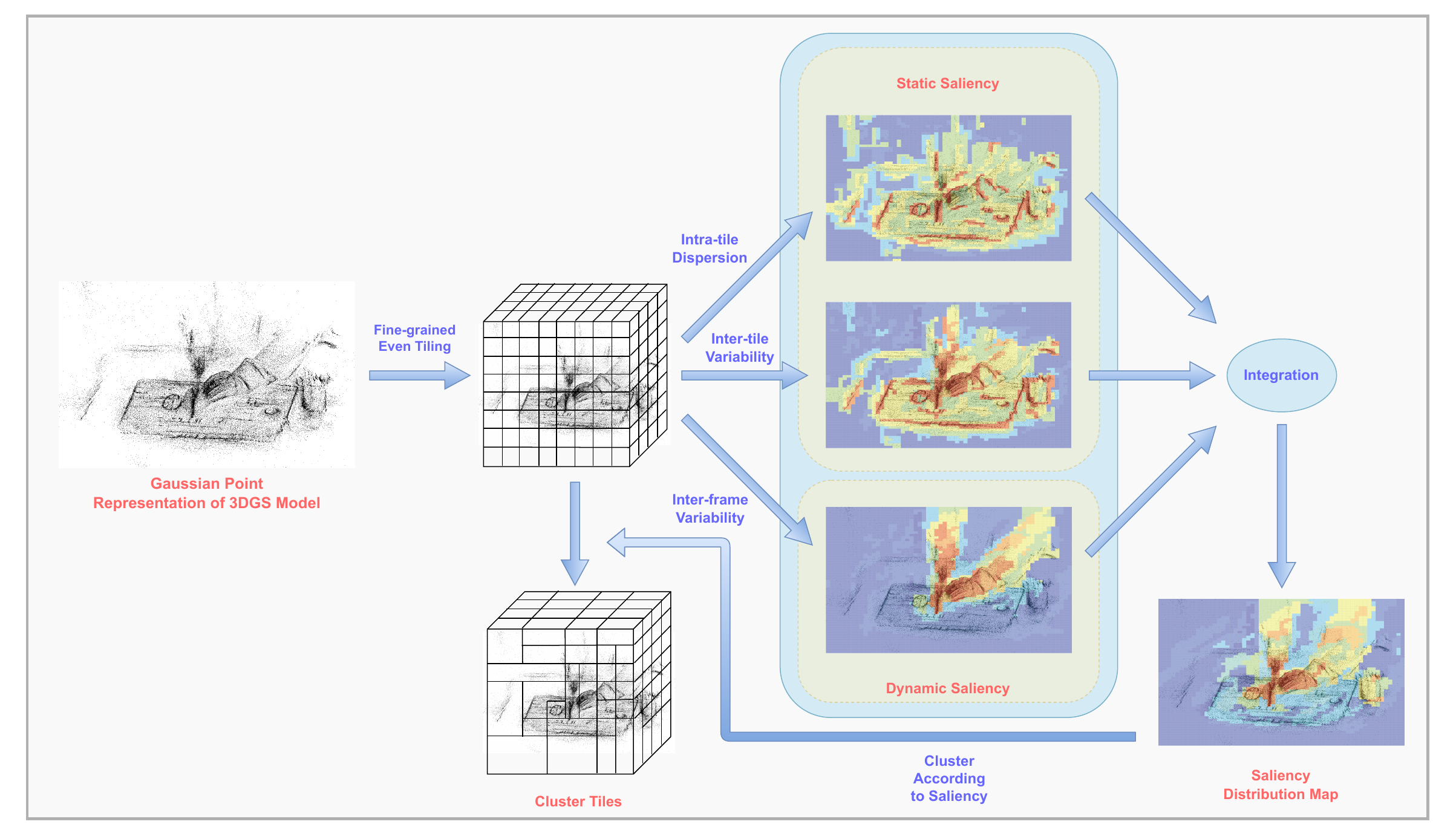}
    \caption{Saliency extraction and adaptive tiling.}
    \label{fig_Tiling}
\end{figure*}

\section{Adaptive 3DGS Video Streaming} 

\subsection{Dynamic 3DGS Video Construction}

3DGS technology is inherently designed for static scene reconstruction. When applied to dynamic scenes, a straightforward approach is to treat each frame as an independent static model. However, this per-frame modeling strategy leads to extremely large data volumes, as each frame independently stores a complete set of Gaussian primitives, resulting in significant redundancy across temporally adjacent frames \cite{wang2024v}. This makes the direct application of static 3DGS techniques to dynamic video both inefficient and impractical. To overcome these limitations, researchers have explored extending 3DGS to incorporate temporal dynamics by constructing models that evolve over time. Nevertheless, such dynamic modeling introduces new challenges, including difficulties in capturing long-range temporal dependencies—often causing motion drift and geometric inconsistencies over extended sequences \cite{xu2024representing}—and limited adaptability to scenes with large or complex motion variations.

To address these issues, we propose a novel method to independently construct Gaussian deformation fields within each GoF. The initial modeling, combined with a deformation field, represents all frames within each GoF. The short duration of each GoF limits the degree of motion within a single segment. This effectively addresses the challenge where Gaussian deformation fields typically struggle to adapt to modeling long sequences and accurately capturing a wide range of motion dynamics. 

Initially, we temporally segment the input dynamic sequence into multiple GoFs and subsequently introduce a dynamic Multi-Layer Perceptron (MLP) with adjustable parameters. These tuning parameters assign weights to each parameter of the MLP during training, thereby directing focus toward the dynamic elements of the scene. This approach enables the decoupling of static and dynamic Gaussian components in a smoothly weighted fashion. We first construct a scene model that captures the overall information across the long sequence, then obtain an initial model for each GoF by preserving extensive static regions of the scene while exclusively fine-tuning the dynamic areas. Through training, we subsequently derive the corresponding deformation field within each GoF. 


Additionally, we perform foreground and background separation for the video content. In cases where the video includes not only the main subject but also the surrounding environment, 3DGS distinctly differentiates between the foreground and background. In a typical 3DGS video, the dynamic changes are primarily observed in the foreground elements, such as the human body, interacting objects, or animals. By separating the video content into foreground and background, we significantly reduce the complexity of the deformation field training. Moreover, during the subsequent streaming process, we can further optimize the transmission by selectively streaming only the foreground model based on the user's interaction demands or rendering the entire scene before transmission.


\begin{figure*}[htb]
    \centering
    \includegraphics[width=1\textwidth]{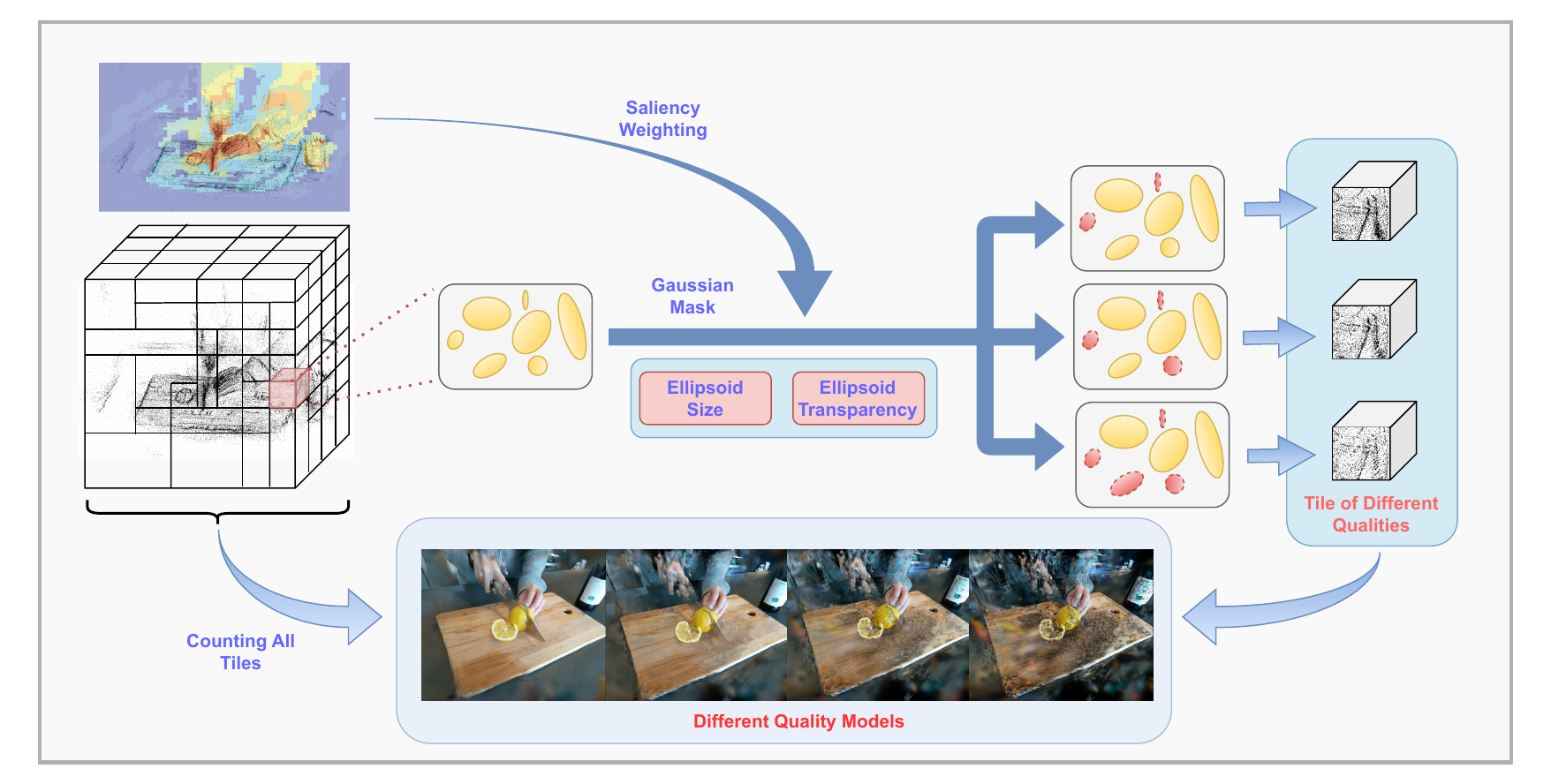}
    \caption{Differentiated quality modeling method.}
    \label{fig_Differentiated}
\end{figure*}

\subsection{Saliency Fusion and Adaptive Tiling of 3DGS Video}

To enable high-quality 3DGS video transmission, we propose an adaptive tiling method based on hybrid saliency detection (Fig. \ref{fig_Tiling}). Each GoF is initially divided into fine-grained uniform tiles. We then extract static saliency, capturing geometric features via intra-tile dispersion ($S^I$) and inter-tile variability ($S^E$), and dynamic saliency ($S^D$), representing motion changes between the first and last frames by comparing matching tiles. Combining these saliency metrics produces a significance distribution map, guiding adaptive tiling through clustering of salient regions.

The weighted fusion of these three saliency metrics yields a comprehensive saliency distribution map for all tiles within a single GoF. Subsequently, a clustering algorithm is employed to group tiles with similar saliency values. Tiles with high saliency and proximal distances are prioritized, merging into larger tiles that represent regions of greater user interest. This adaptive tiling approach ensures that significant areas receive higher quality representation, enhancing the overall viewing experience.

\subsection{Differentiated Quality Modeling of 3DGS Video}




To align our scheme with the practical requirements of various communication scenarios, we propose a differentiated quality modeling approach for 3DGS videos, as illustrated in Fig. \ref{fig_Differentiated}. Unlike point cloud videos, each Gaussian primitive in the 3DGS model carries a unique weight within the overall model, which means we cannot do compression on each Gaussian primitive in the 3DGS model equally. 
We note that Gaussian primitives with lower transparency and smaller ellipsoid sizes have a minimal impact on the overall rendering quality compared to points with higher transparency and larger ellipsoid sizes. Thus, our primary objective is to identify and filter out these less significant Gaussian primitives. For this purpose, we propose a binary Gaussian mask based on significance distribution, which considers both opacity and ellipsoid size of the Gaussian primitives on the rendering results, and provides a better masking effect than considering one of these aspects alone.

\begin{figure*}[!htb]
    \centering

    \begin{subfigure}[t]{0.33\textwidth}
        \centering
        \includegraphics[width=\textwidth]{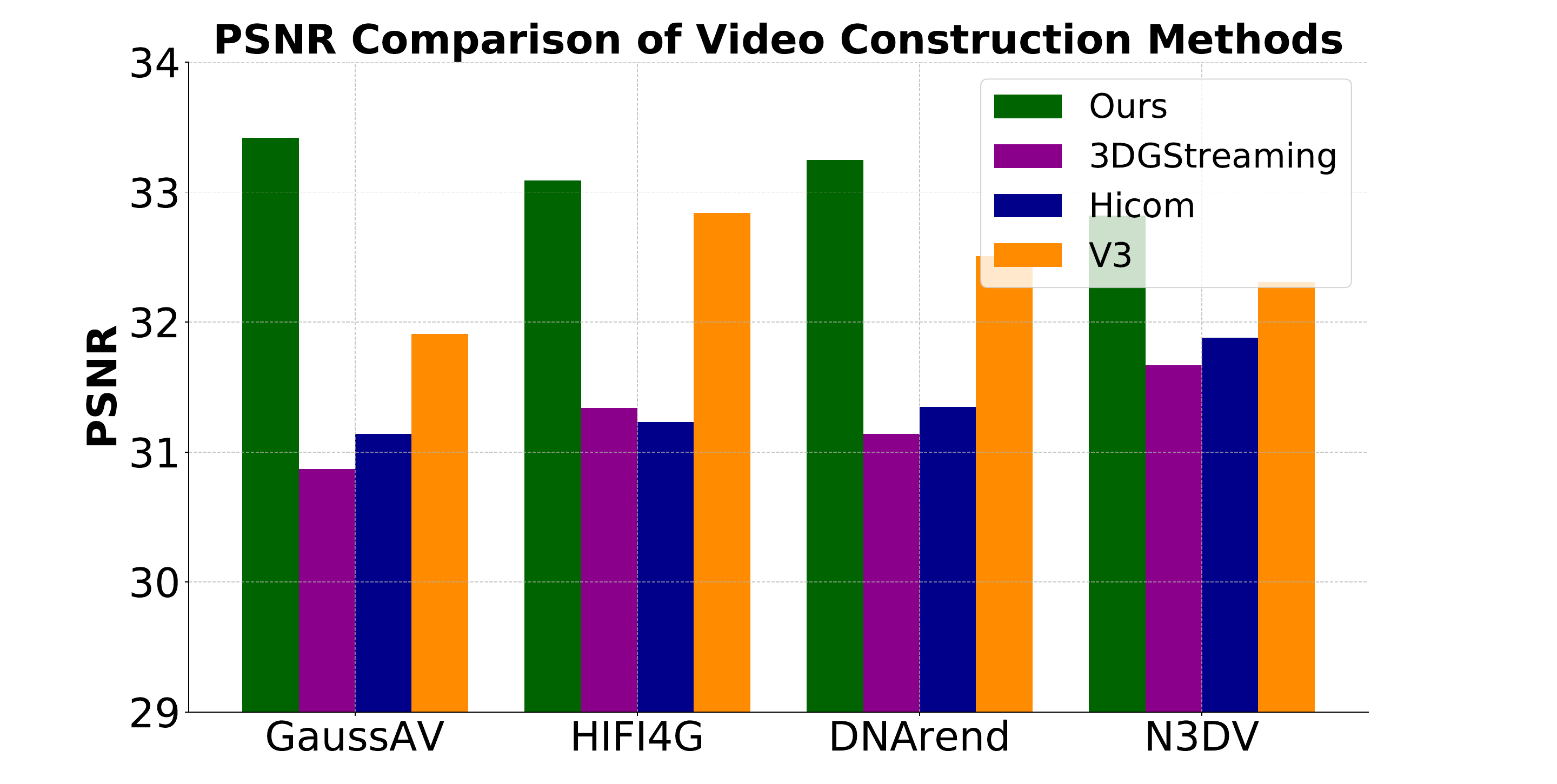}
        \caption{}
        \label{fig:result1}
    \end{subfigure}%
    \hfill
    \begin{subfigure}[t]{0.33\textwidth}
        \centering
        \includegraphics[width=\textwidth]{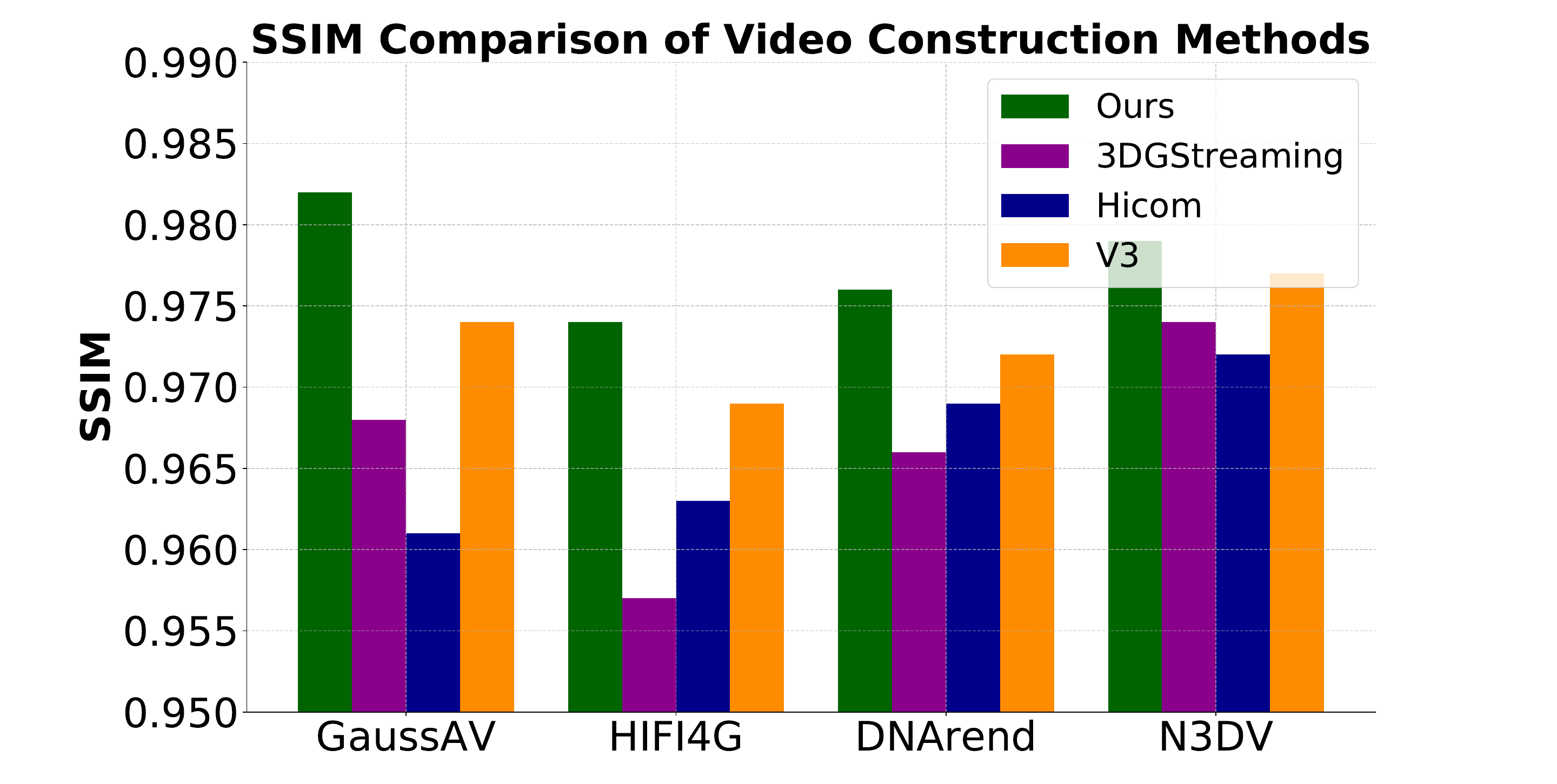}
        \caption{}
        \label{fig:result2}
    \end{subfigure}
    \hfill
    \begin{subfigure}[t]{0.33\textwidth}
        \centering
        \includegraphics[width=\textwidth]{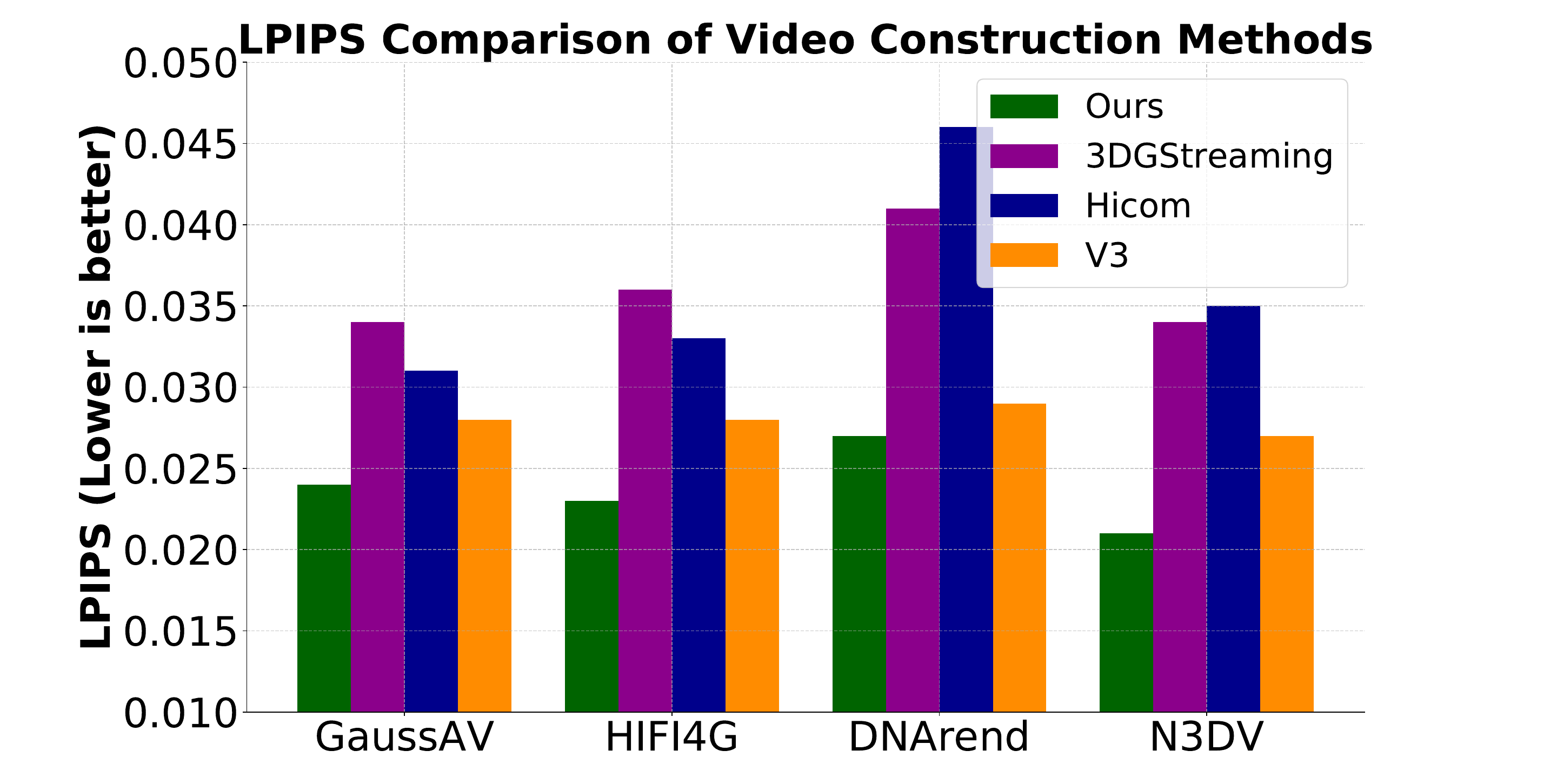}
        \caption{}
        \label{fig:result3}
    \end{subfigure}%
    
    \vspace{0.2cm} 

    \begin{subfigure}[t]{0.33\textwidth}
        \centering
        \includegraphics[width=\textwidth]{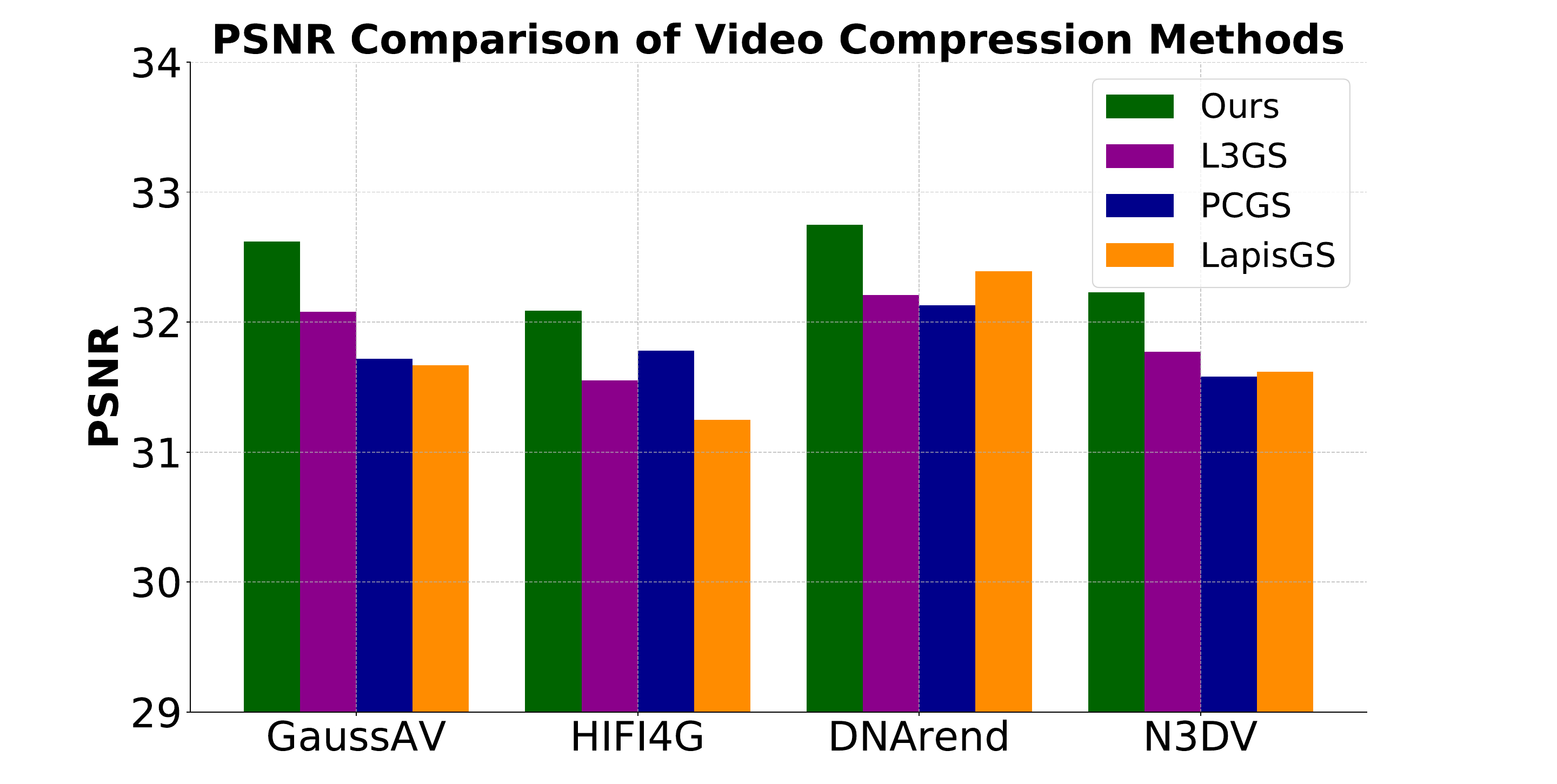}
        \caption{}
        \label{fig:result4}
    \end{subfigure}
    \hfill
    \begin{subfigure}[t]{0.33\textwidth}
        \centering
        \includegraphics[width=\textwidth]{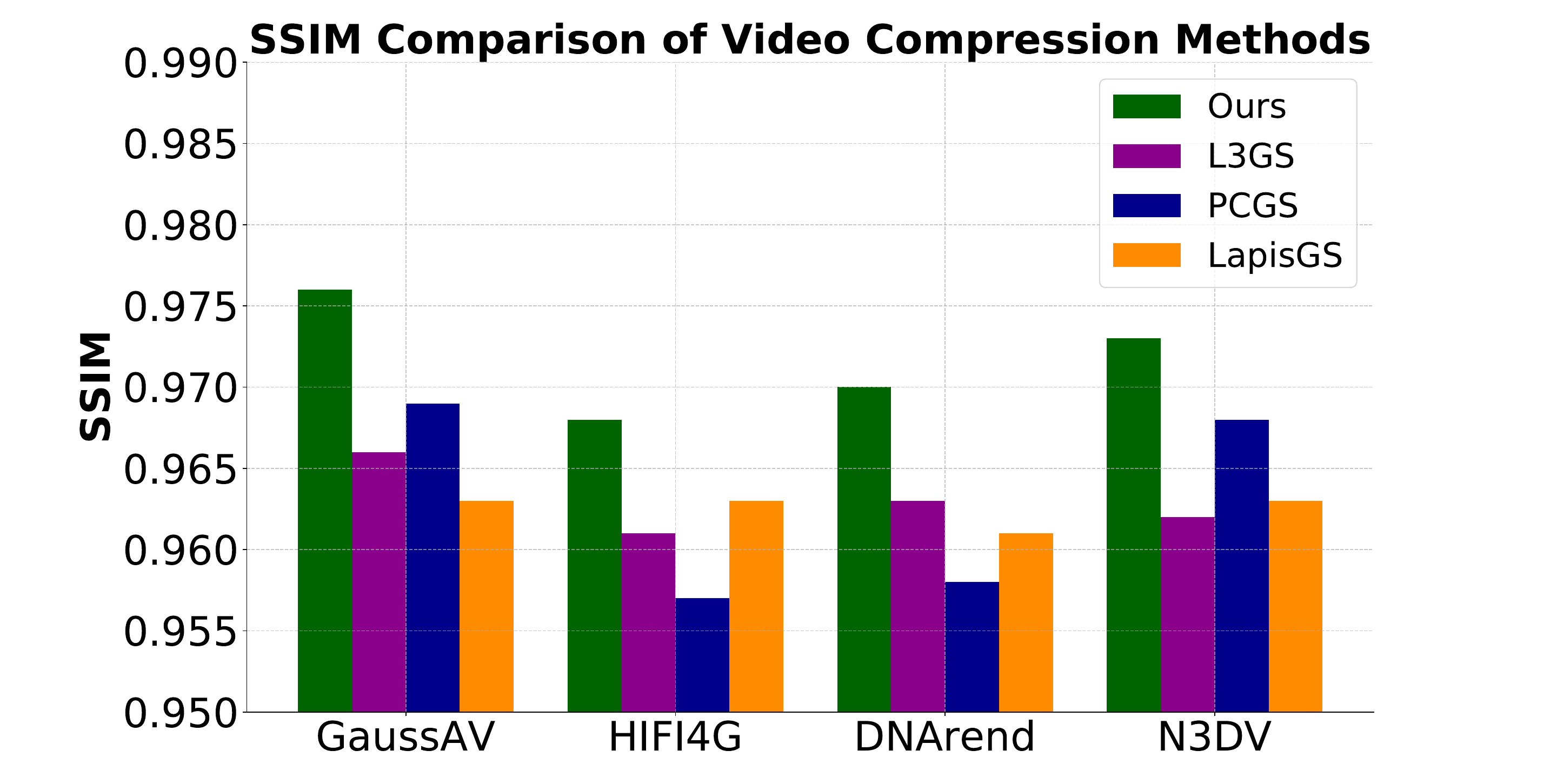}
        \caption{}
        \label{fig:result5}
    \end{subfigure}%
    \hfill
    \begin{subfigure}[t]{0.33\textwidth}
        \centering
        \includegraphics[width=\textwidth]{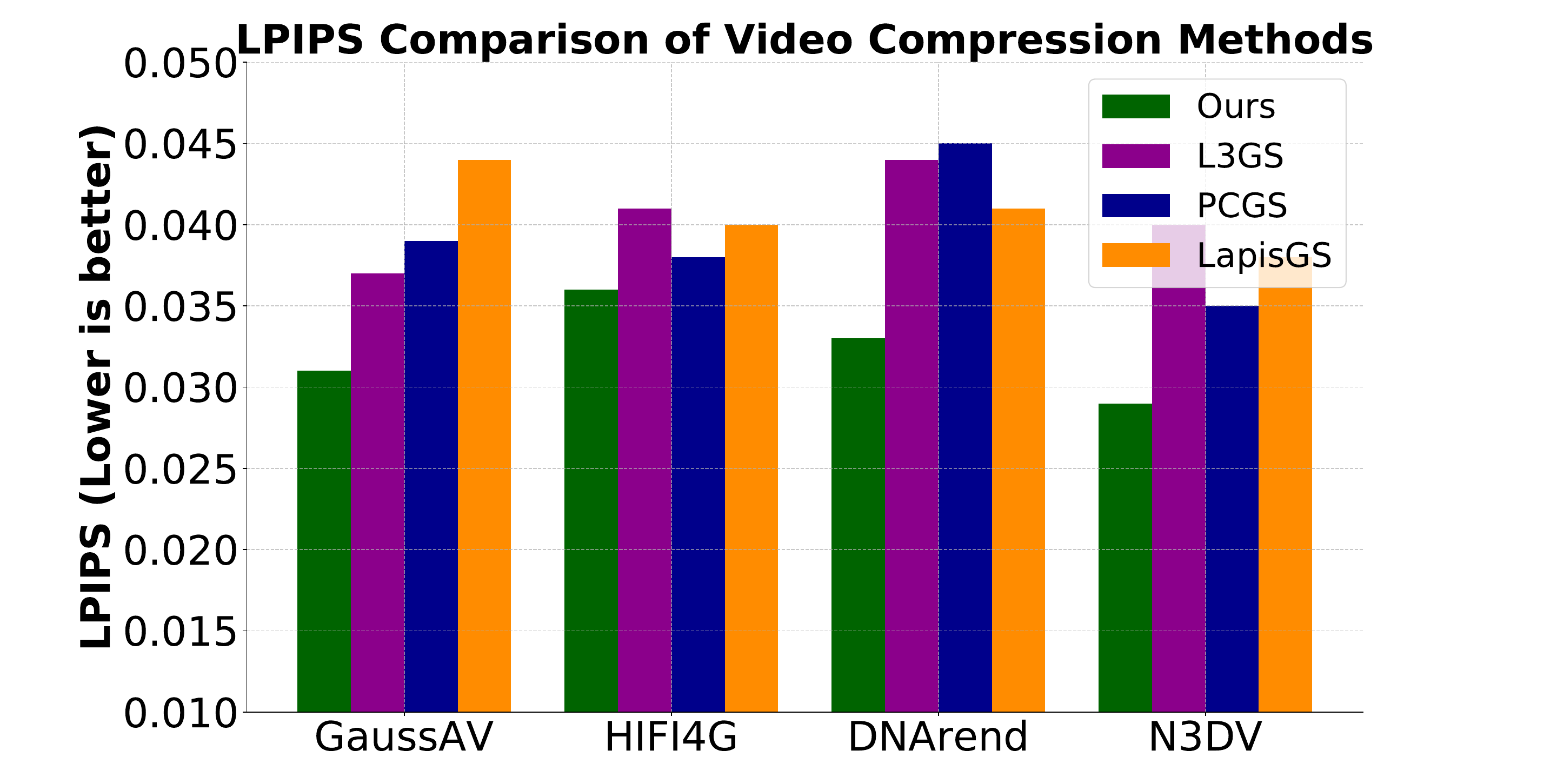}
        \caption{}
        \label{fig:result6}
    \end{subfigure}

    \vspace{0.2cm} 

    \begin{subfigure}[t]{0.33\textwidth}
        \centering
        \includegraphics[width=\textwidth]{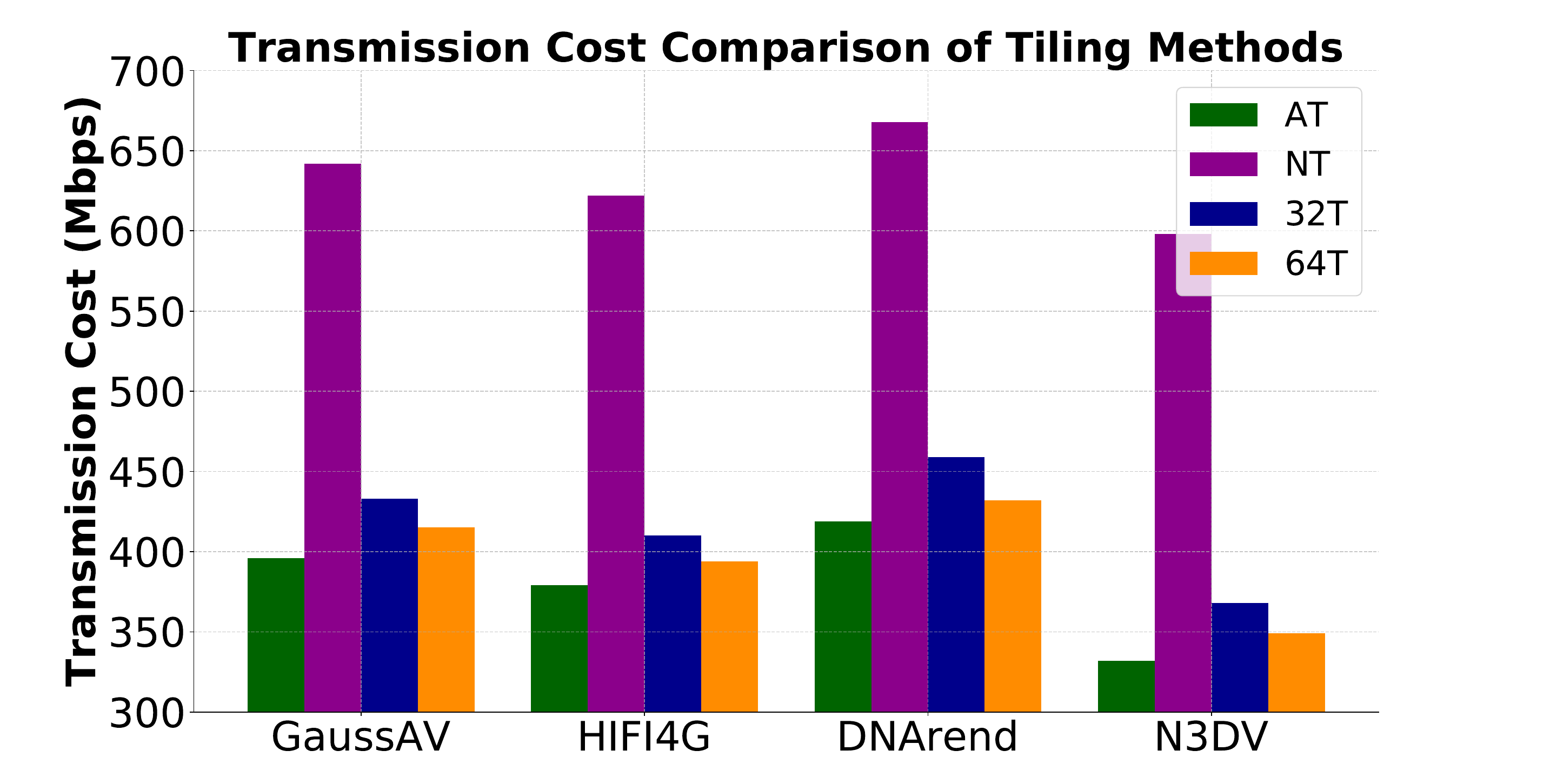}
        \caption{}
        \label{fig:result7}
    \end{subfigure}%
    \hfill
    \begin{subfigure}[t]{0.33\textwidth}
        \centering
        \includegraphics[width=\textwidth]{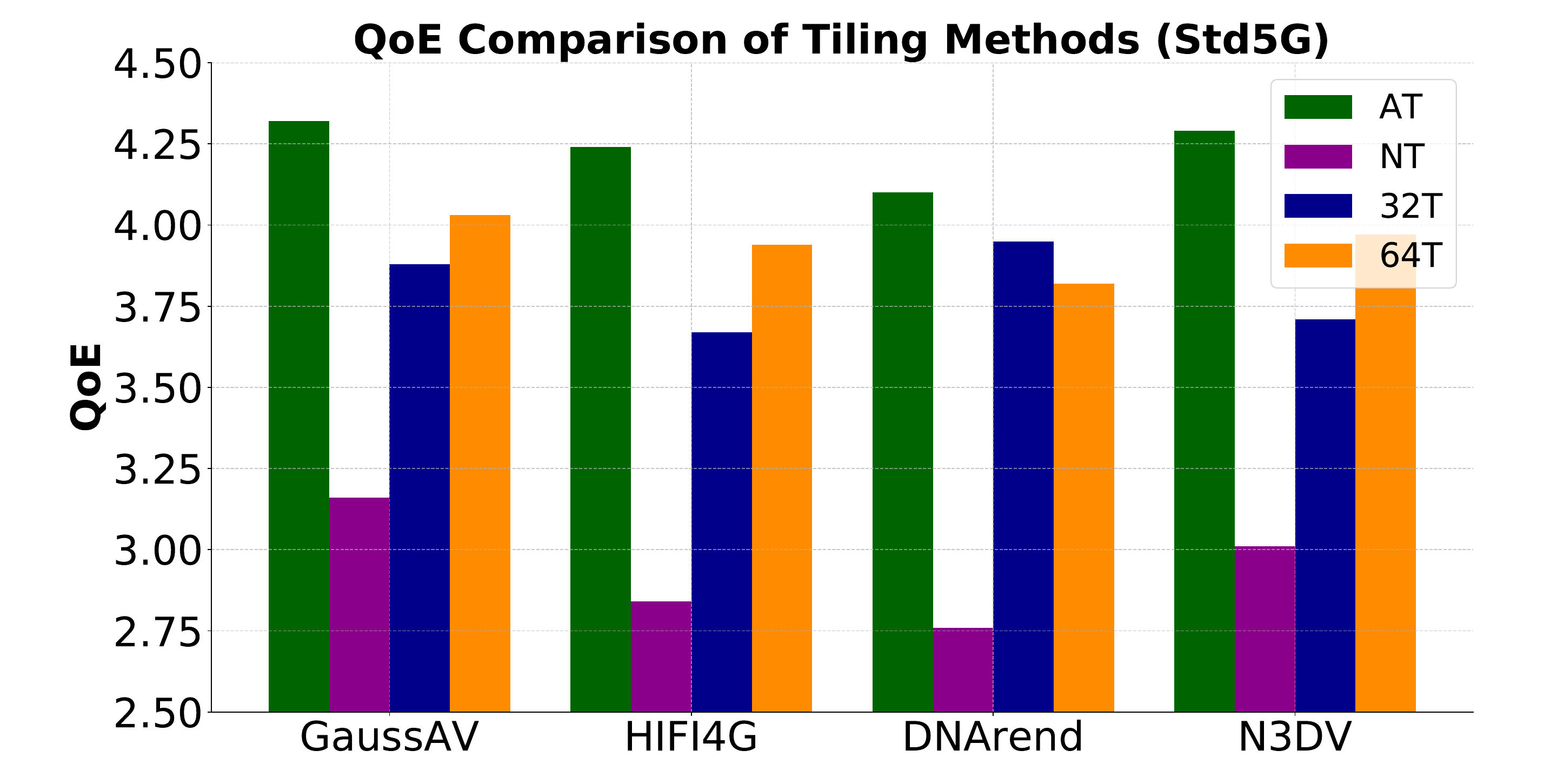}
        \caption{}
        \label{fig:result8}
    \end{subfigure}
    \hfill
    \begin{subfigure}[t]{0.33\textwidth}
        \centering
        \includegraphics[width=\textwidth]{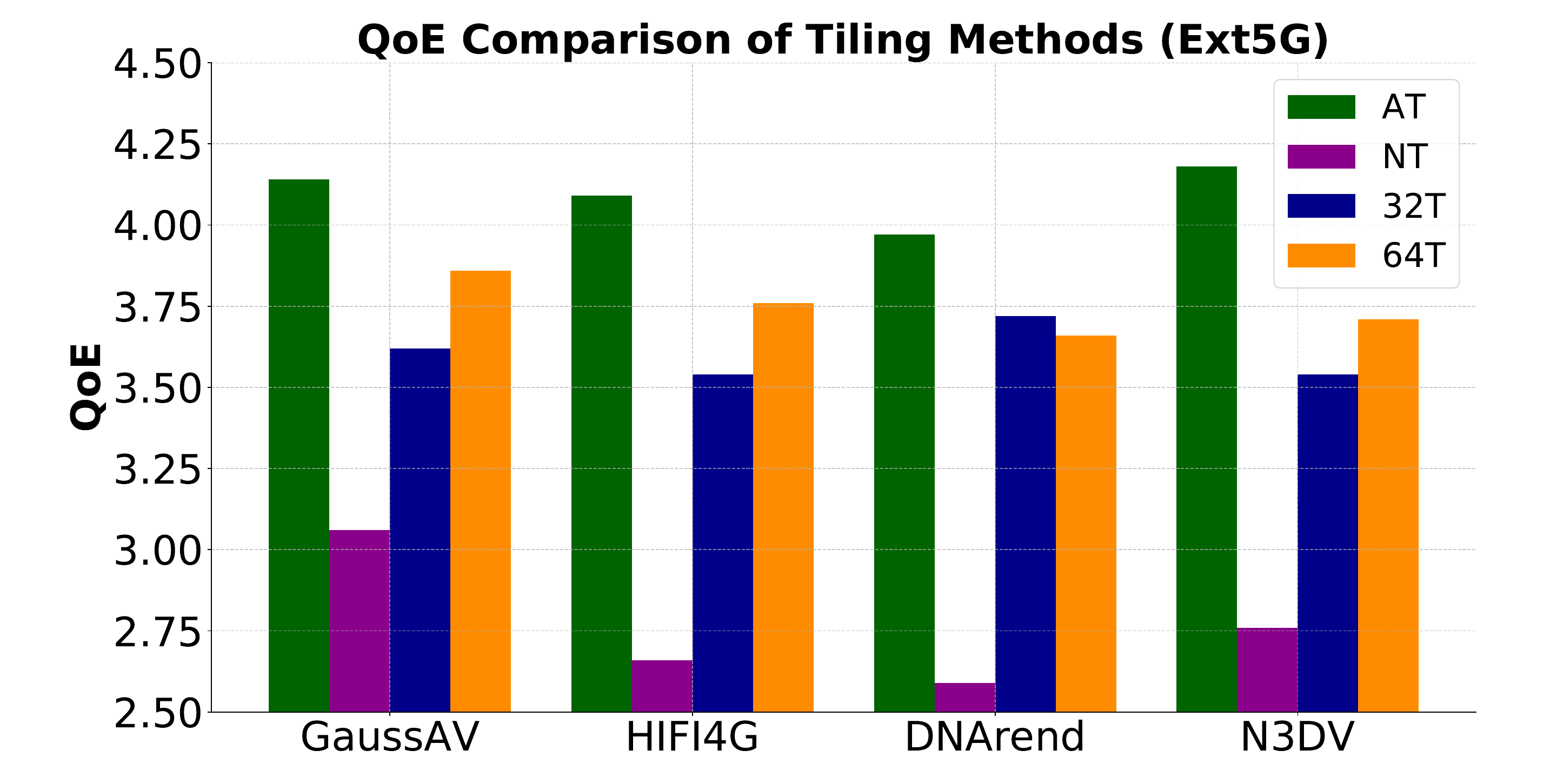}
        \caption{}
        \label{fig:result9}
    \end{subfigure}

    \vspace{0.2cm}
    \begin{subfigure}[t]{0.4\textwidth}
        \centering
        \includegraphics[width=\textwidth]{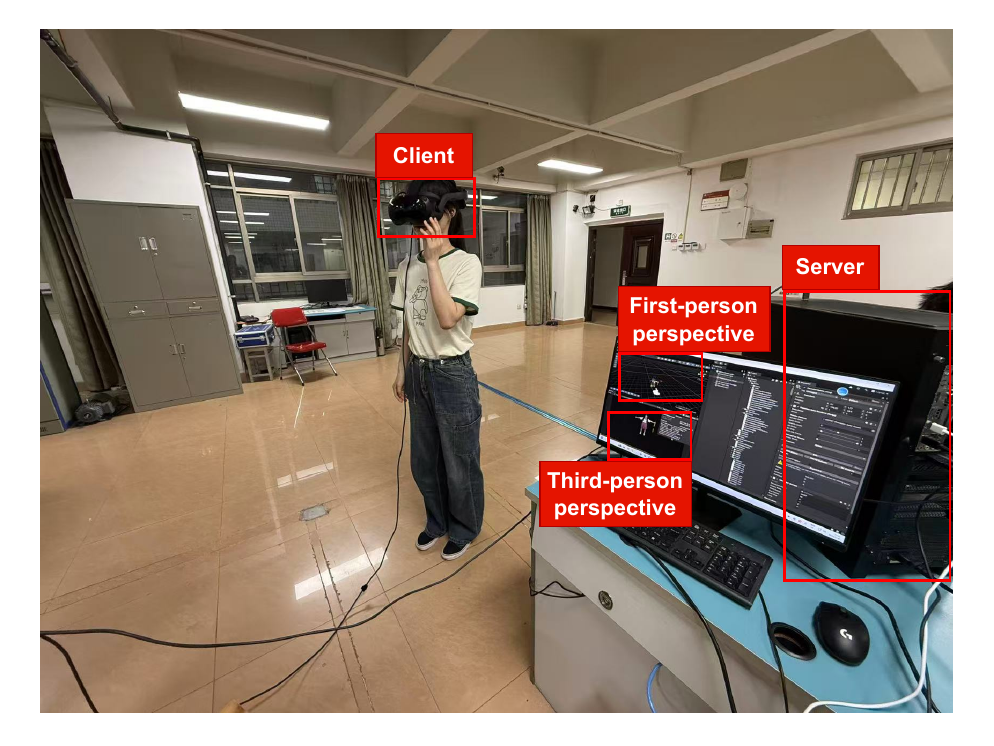}
        \caption{}
        \label{device}
    \end{subfigure}

    \caption{Comparison of the streaming result of different methods.}
    \label{fig_Result}
\end{figure*}

By establishing three distinct masking thresholds, we classified the 3DGS videos within each GoF into four tiers of models with varying qualities.  When determining these thresholds, we referenced the previously obtained saliency distribution maps and adjusted the thresholds based on the saliency weights within different tiles.  For high-saliency regions, we retained as many Gaussian primitives as possible to ensure minimal impact on these visually critical areas.  Conversely, for less visually significant regions, we increased the thresholds to exclude unnecessary Gaussian primitives. For the specific data size of the model for each class, we then refer to the classic bandwidth scenario for the settings. Finally, we encode all the tiles of different quality levels to facilitate the subsequent delivery of the user's desired tiles from the server end to the client end.

\subsection{Adaptive Quality Selection for Tiles}




For the final transmission of 3DGS videos, it is crucial to determine the appropriate quality level for different tiles to optimize transmission efficiency.  To achieve this, we require a robust rating metric that facilitates optimal transmission decisions.  In traditional streaming media, the primary metric for assessing system performance is the user's QoE \cite{liu2021point}. We develop a QoE model tailored for 3DGS videos, drawing from established models used for traditional video.  This model helps us select the specific quality levels of tiles by maximizing the overall QoE.

For a single GoF, its QoE is primarily influenced by three factors: the average video quality, the total stall time, and the stall frequency. Both excessive stall duration and the frequency of buffering events adversely affect the user experience, making QoE inversely proportional to these two factors. The final QoE model is defined as a weighted sum of these three factors. For the average video quality assessment, we extend the standard 2D image-based fidelity metrics used in 3DGS by incorporating an additional geometric-space distortion metric, thereby constructing a unified video quality representation. Specifically, for each GoF, the geometric consistency between the highest-quality tile representation and the current tile selection is evaluated via bidirectional nearest-neighbor matching, following an iterative closest point alignment. For each matched 3D Gaussian pair, geometric distortion is quantified through spatial deviations, while color fidelity is assessed by comparing zeroth-order spherical harmonics coefficients. These spatial and chromatic errors are integrated into a comprehensive PSNR metric with adaptive sampling compensation based on point density. This enhanced evaluation more accurately reflects perceptual quality by jointly considering both 2D rendering fidelity and underlying 3D structural differences.

Requesting tiles from the client to server is modeled as an optimization problem to maximize QoE under bandwidth constraints. Using user FoV and bandwidth predictions, an integer linear programming (ILP) solver selects the optimal subset of tiles and quality levels. The server then transmits these tiles and deformation data to the client, which decodes and assembles them into a complete 3DGS model for playback from a buffer, ensuring smooth video rendering.

\section{Experimental Results and Analysis}

\begin{figure*}[htb]
    \centering
    \includegraphics[width=0.8\textwidth]{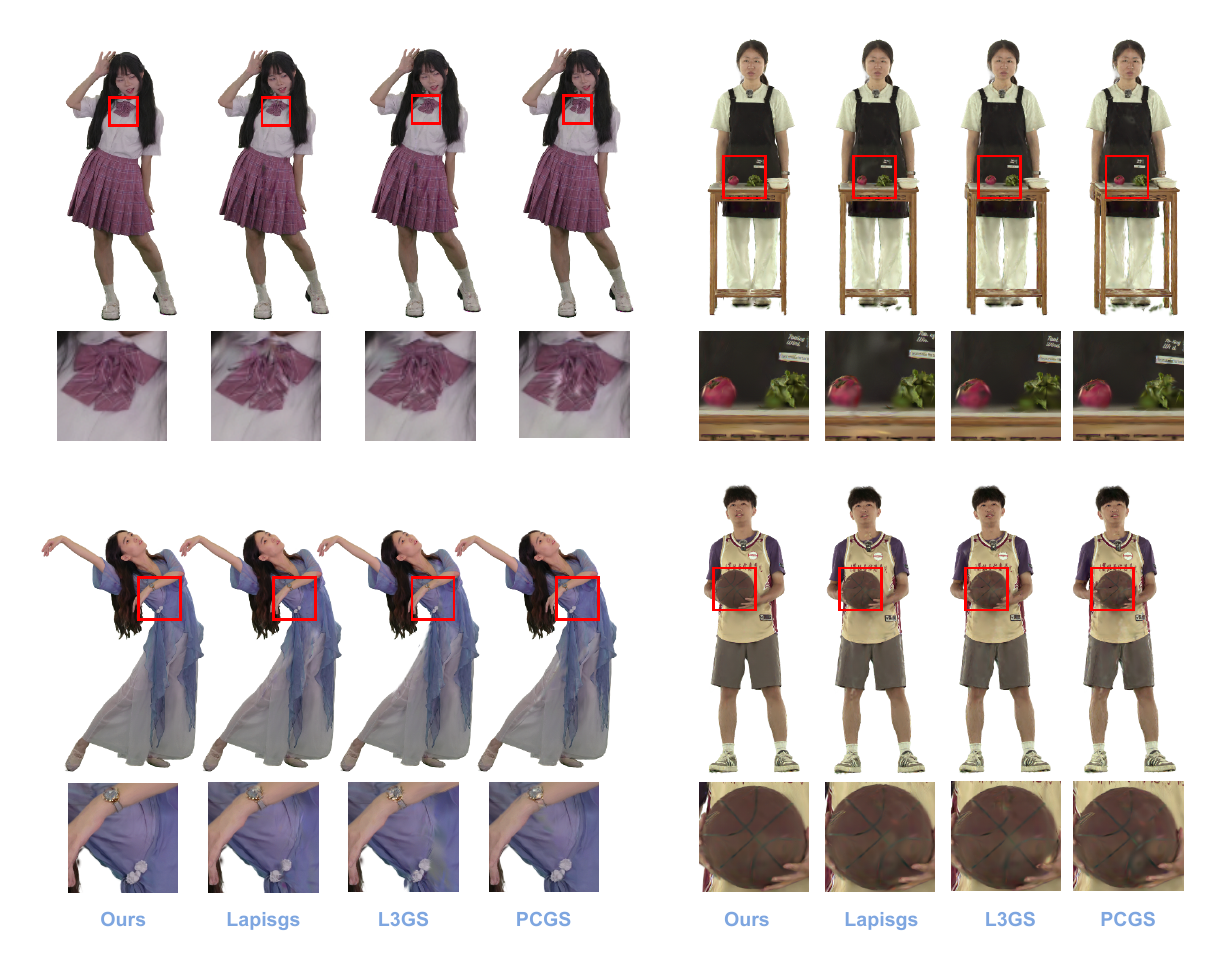}
    \caption{Qualitative comparison across multiple scenes under the 70\% compression ratio.}
    \label{fig_qualitative_results}
\end{figure*}

In this section, we conduct a comprehensive evaluation of our framework by comparing it with current state-of-the-art methods across three critical dimensions: video construction, compression strategy, and tiling scheme. Our experiments are conducted on four datasets—N3DV \cite{li2022neural}, HIFI4G \cite{jiang2024hifi4g}, DNA-Rendering \cite{cheng2023dna}, and our self-recorded GaussAV—each selected to represent different scene types and video characteristics. Specifically, N3DV typically consists of small-scale indoor environments with constrained front-facing 180-degree camera trajectories, reflecting controlled capture settings; DNA-Rendering features large-scale scenes with complex and unconstrained backgrounds; HIFI4G focuses on single-person scenarios with intricate motion patterns; and GaussAV introduces multi-person interactions, posing challenges related to occlusion and inter-object relationships. To assess the adaptability of our framework under varying network conditions, we also simulate diverse 5G bandwidth environments, evaluating its performance in terms of visual quality, transmission cost, and user-perceived quality (QoE).

\begin{itemize}
   
\item \textbf{Video construction method}: To evaluate the effectiveness of our video construction strategy, we compare it against three recent state-of-the-art methods targeting streamable 3DGS video: 3DGStream \cite{sun20243dgstream}, HiCoM \cite{gao2024hicom}, and V3 \cite{wang2024v}. These methods focus on real-time reconstruction and rendering of dynamic 3D Gaussian content. Specifically, 3DGStream employs frame-by-frame online optimization with neural transformation caching; HiCoM introduces a hierarchical coherent motion model and persistent optimization strategy; and V3 aims to support 3DGS streaming across diverse platforms. We evaluate video quality using three commonly adopted metrics: PSNR, SSIM, and LPIPS. As shown in Fig. \ref{fig:result1}–\ref{fig:result3}, our method consistently achieves superior reconstruction performance across all sequences. This improvement stems from our segmented GoF reconstruction approach, which effectively mitigates deformation field drift over time and enhances temporal stability in dynamic scenes.

\item \textbf{Video compression method}: To evaluate the effectiveness of our differential quality modeling scheme in the context of 3DGS video compression, we compare our method against three recently proposed state-of-the-art approaches: L3GS \cite{tsai2025l3gs}, PCGS \cite{chen2025pcgs}, and LapisGS \cite{shi2024lapisgs}. All three methods adopt hierarchical or progressive strategies to enable resource-adaptive 3DGS compression, conceptually aligning with our quality stratification approach. We conduct a fair comparison by applying a consistent 70\% compression ratio across all methods and evaluate the reconstructed video quality using PSNR, SSIM, and LPIPS metrics. As shown in Fig. \ref{fig:result4}–\ref{fig:result6}, our method achieves the highest reconstruction quality on all datasets. This advantage is primarily attributed to our Gaussian mask-based compression strategy, which effectively preserves visually salient regions and foreground structures while aggressively eliminating less important primitives, thereby maximizing compression efficiency without sacrificing perceptual quality. In addition to quantitative metrics, we further supplement our analysis with qualitative results across multiple sequences. As shown in Fig. \ref{fig_qualitative_results}, visual comparisons demonstrate that our approach achieves clearer structure, better detail retention, and fewer visual artifacts compared to other SOTA methods—validating the practical benefits of our compression scheme in perceptual quality.


\item \textbf{Tiling method}: To validate the transmission superiority of our adaptive tiling scheme (AT), we conducted comprehensive comparisons against three baseline approaches—NT (No Tiling), 64T (64 uniform tiles), and 32T (32 uniform tiles)—under two distinct 5G network environments: a Standard 5G environment (Std5G) with stable bandwidth (350-700 Mbps) and an Extreme 5G environment (Ext5G) exhibiting highly erratic bandwidth patterns (0-1200 Mbps). Transmission cost analysis under Std5G conditions (Fig. \ref{fig:result7}) confirms AT consistently achieves the lowest data volume across all datasets. Furthermore, QoE evaluations demonstrate AT's robust performance: in Std5G (Fig. \ref{fig:result8}), AT outperforms all baselines by optimally balancing tile granularity and coverage area; in Ext5G (Fig. \ref{fig:result9}), it maintains superior QoE by dynamically adapting to bandwidth fluctuations. This dual-environment validation confirms AT effectively resolves core tiling tradeoffs: preventing excessive decoding latency from fine-grained tiling (64T) while eliminating FoV-overflow redundancy in coarse tiling schemes (NT/32T).

\item \textbf{Computation overhead}: Through comprehensive benchmarking on our experimental platforms—a desktop system equipped with GeForce RTX 4080 GPU and Intel i5-14600KF CPU, alongside the Meta Quest 3 head-mounted XR device (Fig. \ref{device}). Our framework achieves an average training time of 58.3 seconds per frame. Rendering performance reaches 246 FPS on desktop platforms and 126 FPS on Meta Quest 3, confirming real-time six-degree-of-freedom viewpoint rendering capability. Regarding decoding efficiency: simple scenes (e.g., small-scale N3DV scenarios or single-person HIFI4G sequences) achieve approximately 20 ms per-frame decoding latency, whereas complex scenes (GaussAV multi-person environments or large-scale DNA-Rendering scenarios) consistently exceed 33 ms per frame—surpassing the 30FPS real-time threshold and introducing playback delays. To resolve this bottleneck, we implement a dual-mode transmission scheme comprising encoded tiles (compressed via deformation fields and client-decoded post-transmission) and reconstructed tiles (pre-reconstructed and directly transmitted), enabling QoE-driven dynamic selection where bandwidth abundance allows prioritized transmission of reconstructed tiles to alleviate client-side computational pressure, thereby ensuring end-to-end real-time 3DGS video streaming across all tested scenarios.

\end{itemize}

\section{Open Research Issues and Outlook}





In this paper, while we have conducted preliminary investigations into 3DGS video transmission, the intersection of 3DGS technology and volumetric video represents a nascent research domain fraught with numerous unresolved issues. We will outline these challenges below and propose potential solutions to address them.

\textbf{Video flickering between neighboring GoFs}: Since the deformation field cannot perfectly represent the actual scene motion, using the initial model and the corresponding deformation field representation according to different GoFs will result in a large difference between the last frame of each GoF and the first frame of the next GoF, and this difference will cause flickering between GoFs during video playback. One possible solution to this problem is to insert buffer frames between GoFs to create a smooth transition.



\textbf{Cross-platform compatibility}: While our current framework supports real-time rendering on desktop systems and XR devices, extending support to mobile platforms remains an open challenge. Mobile devices typically lack the storage and computational resources required for full-resolution 3DGS model rendering. As such, achieving cross-platform compatibility will require developing lightweight decoders and optimizing the transmission pipeline for mobile SoCs. We identify this as a key direction for future work, aiming to enable efficient and high-quality 3DGS streaming across a broader range of devices.

\section{Conclusions}

In this paper, we proposed a pioneering framework for 3DGS video streaming. Our framework encompassed the entire process from constructing 3DGS volumetric videos to achieving optimal transmission to clients. To validate our proposed scheme, we implemented a transmission system based on these ideas and conducted experiments to assess its feasibility in real-world scenarios. The experimental results demonstrated that our method exhibited advantages in streaming 3DGS videos, showing superior performance in video quality, compression effectiveness, and transmission rate.






\ifCLASSOPTIONcaptionsoff
  \newpage
\fi





\bibliographystyle{IEEEtran}


\end{document}